\documentclass[10pt,twocolumn,letterpaper]{article}

\usepackage[pagenumbers]{cvpr} % for an arXiv version
\usepackage[pagebackref,breaklinks,colorlinks,citecolor=cvprblue]{hyperref}
\usepackage{booktabs}
\usepackage{multirow}
\usepackage{balance}
\usepackage[dvipsnames]{xcolor}
\definecolor{cvprblue}{rgb}{0.21,0.49,0.74}
\definecolor{myblue}{RGB}{46, 105, 167}
\definecolor{myorange}{RGB}{186, 82, 30}
\definecolor{ForestGreen}{rgb}{0.13, 0.55, 0.13}
\definecolor{Maroon}{rgb}{0.69, 0.19, 0.0}
\definecolor{my_pink}{rgb}{1,0.43,0.41}
\definecolor{my_gray}{rgb}{0,0,0.2}
\definecolor{citecolor}{RGB}{30,130,255}
\definecolor{LightCyan}{rgb}{0.88,1,1}
\def\etal{\emph{et al. }}

\def\eg{\emph{e.g., }}
\newcommand\mypara[1]{\vspace{1mm}\noindent\textbf{#1}}
\newcommand{\nocontentsline}[3]{}
\newcommand{\tocless}[2]{\bgroup\let\addcontentsline=\nocontentsline#1{#2}\egroup}

\def\paperID{9715} 
\def\confName{CVPR}
\def\confYear{2024}

\title{Towards Effective Usage of Human-Centric Priors in Diffusion Models for Text-based Human Image Generation}

\author{
Junyan Wang $^{1}$ \quad
Zhenhong Sun $^{2}$ \quad
Zhiyu Tan $^{3}$ \quad
Xuanbai Chen $^{4}$ \quad 
Weihua Chen $^{5}$ \\ [2pt]\quad 
Hao Li $^{6}$\thanks{Corresponding author}  \quad
Cheng Zhang $^{4}$ \quad
Yang Song $^{1}$\\ [6pt]
{
$^{1}$ {University of New South Wales} \quad
$^{2}$ {Australian National University} \quad
$^{3}$ {INF Technology}
}  \\
{
$^{4}$ {Carnegie Mellon University} \quad
$^{5}$ {Alibaba DAMO Academy} \quad
$^{6}$ {Fudan University}
}
}

\begin{document}
\maketitle
\begin{abstract}
\label{sec:abs}
Vanilla text-to-image diffusion models struggle with generating accurate human images, commonly resulting in imperfect anatomies such as unnatural postures or disproportionate limbs.
Existing methods address this issue mostly by fine-tuning the model with extra images or adding additional controls --- human-centric priors such as pose or depth maps --- during the image generation phase. This paper explores the integration of these human-centric priors directly into the model fine-tuning stage, essentially eliminating the need for extra conditions at the inference stage. We realize this idea by proposing a human-centric alignment loss to strengthen human-related information from the textual prompts within the cross-attention maps. To ensure semantic detail richness and human structural accuracy during fine-tuning, we introduce scale-aware and step-wise constraints within the diffusion process, according to an in-depth analysis of the cross-attention layer. Extensive experiments show that our method largely improves over state-of-the-art text-to-image models to synthesize high-quality human images based on user-written prompts. Project page: \url{https://hcplayercvpr2024.github.io}. 

\end{abstract}

\tocless
\section{Introduction}
\label{sec:intro}

\begin{figure}
\centering
    \includegraphics[width=0.8\linewidth]{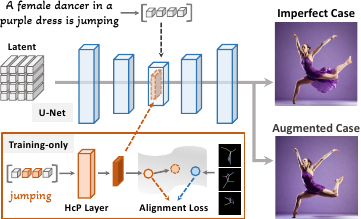}\vspace{-1mm}
    \caption{\small Existing text-to-image models often struggle to generate human images with accurate anatomy (upper branch). We incorporate human-centric priors into the model fine-tuning stage to rectify this imperfection (bottom branch). The learned model can synthesize high-quality human images from text without requiring additional conditions at the inference stage.}
    \label{fig:intro}\vspace{-2.5mm}
\end{figure}

Recent advancements in diffusion models have significantly improved \textbf{text-to-image} (T2I) generation, consistently enhancing the quality and precision of visual synthesis from textual descriptions \cite{ramesh2021zero,tao2022df,podell2023sdxl,saharia2022photorealistic}. 
Within the paradigm of T2I, generating human images emerges as a specific focus, drawing substantial attention for its potential in applications such as virtual try-on \cite{zhou2022cross} and entertainment \cite{pan2022synthesizing}.
Despite the remarkable advancements, human image generation still faces challenges, including the incomplete rendering of the human body, inaccuracies in the portrayal, and limb disproportions, such as the imperfect case shown in Figure \ref{fig:intro}.
The challenges in generating human images arise from the diffusion model's inherent emphasis on broad generalization across diverse data, leading to a lack of detailed attention to human structure in the generated images.
Resolving these issues is essential for advancing the field toward producing more realistic and accurate human images from textual descriptions.

The straightforward method to tackle the challenges in human image generation involves using additional conditions during both the training and inference phases, \eg ControlNet \cite{zhang2023adding}. 
While employing extra conditions like pose image guidance indeed improves the structural integrity of human, their reliance on additional conditions does not address the challenges inherent in human image generation. It restricts the direct creation of diverse images from text prompts, and requires extra conditions beyond text during inference, making the process tedious and less end-user friendly.
Alternatively, another efficient approach employs fine-tuning methods, \eg LoRA \cite{hu2021lora},
which adjust pre-trained models on specialized human-centric datasets for more accurate human feature representation. While this approach can enhance human image generation, it may modify the model's original expressiveness and lead to catastrophic forgetting, resulting in outputs that are limited by the characteristics of the fine-tuning dataset.

Thus, our work concentrates on \textbf{text-based Human Image Generation} (tHIG), which relies exclusively on textual inputs without requiring additional conditions during inference. 
The primary objective of tHIG is to address the challenges in human image generation within diffusion models, enhancing their expressive power while leveraging the inherent diversity and simplicity of diffusion models to generate human images without additional conditions.
To tackle the challenges in human image generation, we delved into several key factors for influencing the final output. Firstly, echoing the findings from \cite{hertz2022prompt}, our analysis shows the role of \textit{cross-attention maps} within diffusion models is a fundamental element, significantly impacting the structural content. This impact is particularly crucial in the generation of human body structures, where accurate representation depends critically on these maps' effectiveness.
Furthermore, incorporating \textit{human-centric priors}, such as pose image, has been shown to enhance human representation in synthesized visuals \cite{ju2023humansd}. Aligning this with the inherent capabilities of existing T2I models provides a solid foundation for generating more realistic human figures.

Building on the outlined motivations, our work introduces a novel plug-and-play method for tHIG, which emerges from comprehensive insights into the diffusion process, with a particular focus on the crucial role of \textit{cross-attention maps}.
We present the innovative \textbf{H}uman-\textbf{c}entric \textbf{P}rior (HcP) layer, designed to enhance the alignment between the cross-attention maps and human-centric textual information in the prompt.
Incorporating a specialized Human-centric Alignment loss, the HcP layer effectively integrates other auxiliary human-centric prior information, such as key poses, exclusively during the training phase. 
This inclusion improves the capability of the diffusion model to produce accurate human structure only with textual prompts, without requiring additional conditions during inference.
Furthermore, our approach adopts a step and scale aware training strategy, guided by our in-depth analysis of the cross-attention layers.
This strategy effectively balances the structural accuracy and detail richness in generated human images, while preserving the diversity and creativity inherent to T2I models.

We validate our HcP layer with Stable Diffusion \cite{rombach2022high}.
The HcP layer can preserve the original generative capabilities of the diffusion model and produce high-quality human image generation without requiring additional conditions during the inference phase.
Moreover, the HcP layer is compatible with the existing controllable T2I diffusion models (\eg ControlNet \cite{zhang2023adding}) in a plug-and-play manner.

\vspace{3mm}
\tocless
\section{Related Work}
\label{sec:related}

\mypara{Text-to-Image Generation}.
T2I as a rapidly evolving field, has witnessed the emergence of numerous model architectures and learning paradigms \cite{mansimov2015generating,reed2016generative,ramesh2021zero,tao2022df,yu2022scaling,chang2023muse,ding2022cogview2,yang2022diffusion,croitoru2023diffusion,dhariwal2021diffusion,kingma2021variational,xu2024survey,zhang2023patch,zhang2022contextual,zhang2023patch2,zhang2023inclusive}.
Generative Adversarial Networks (GANs) based models \cite{xu2018attngan,qiao2019mirrorgan,zhu2019dm} initially played a pivotal role in this field, establishing key benchmarks for quality and diversity in generated images.
Recent advancements \cite{rombach2022high,podell2023sdxl,ramesh2021zero,saharia2022photorealistic} in diffusion models have significantly enhanced the capabilities of text-to-image generation.
Diffusion models derive their effectiveness from a structured denoising process \cite{ho2020denoising}, which transforms random noise into coherent images guided by textual descriptions.
For example, latent diffusion \cite{rombach2022high} utilizes a latent space-based approach where it first converts text into a latent representation, which is then progressively refined into detailed images through a denoising process.
In this work, we build upon these diffusion model advancements by introducing HcP layer, specifically designed to enhance HIG.

\mypara{Human Image Synthesis}.
Human image synthesis is an area of significant interest due to its broad applications in industries such as fashion \cite{han2022fashionvil,han2023fame} and entertainment \cite{pan2022synthesizing}. 
Most efforts \cite{ren2020deep,yang2021towards,zhang2021pise,ren2022neural,zhang2023adding,mou2023t2i,ju2023humansd,liu2023hyperhuman,zhang2022exploring,chen2023beyond} to address the challenges of diffusion models in accurately representing human structure have relied on introducing additional conditions during both training and inference stages.
For example, HumanSD \cite{ju2023humansd} proposes a native skeleton-guided diffusion model for controllable human image generation by using a heatmap-guided denoising loss.
However, this approach often complicates the image generation process and can limit the diversity of output images.
Our work introduces the HcP layer and employs a targeted training strategy that enhances human image synthesis in diffusion models without additional conditions,
which ensures the high-quality generation of human images.

\mypara{Image Editing via Cross-Attention}.
Recent advancements in text-driven image editing have shown significant progress, especially within the paradigm of diffusion-based models \cite{esser2021taming,avrahami2022blended,kim2022diffusionclip,liu2023cones}. 
Kim \etal \cite{kim2022diffusionclip} show how to perform global changes, whereas Avrahami \etal \cite{avrahami2022blended} successfully perform local manipulations using user-provided masks for guidance.
Progress in text-driven image editing primarily relies on refining the cross-attention layers within U-Net architectures \cite{hertz2022prompt,ye2023ip,feng2022training}.
For example, the work of Hertz \etal \cite{hertz2022prompt} presents several applications which monitor the image synthesis by editing the textual prompt only. This includes localized editing by replacing a word, global editing by adding a specification, and even delicately controlling the extent to which a word is reflected in the image.
However, our approach enhances the influence of certain text embeddings during image generation, ensuring efficiency without additional conditions at inference.

\tocless
\section{The Proposed Approach}
\label{sec:method}
\begin{figure*}
\centering
    \includegraphics[width=0.93\textwidth]{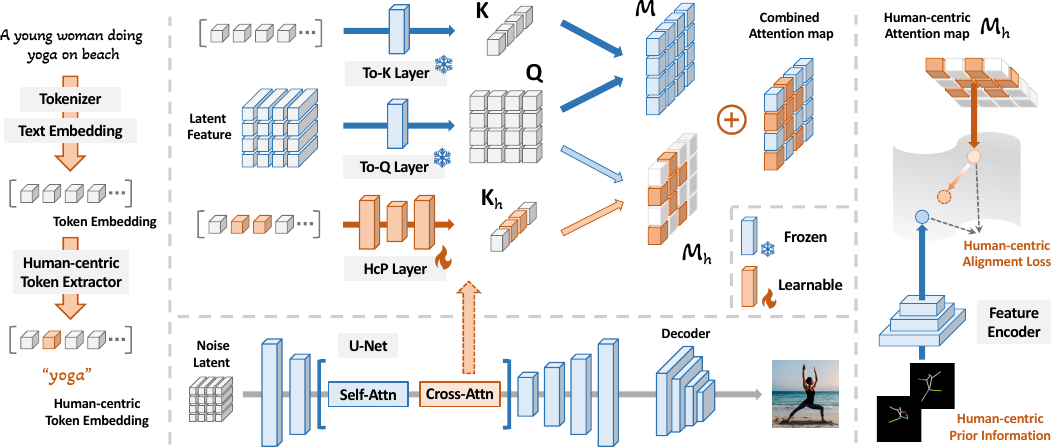}
    \caption{Overview of the proposed \textbf{\textcolor{myorange}{learnable}} Human-centric Prior layer training in the \textbf{\textcolor{myblue}{frozen}} pre-trained latent diffusion model. The left part shows the process of human-centric text tokens extraction, the middle part indicates the overall process of the HcP layer plugged into the U-Net framework, and the right part shows the HcP layer training with the proposed human-centric alignment loss.}
    \label{fig:framework}
\end{figure*}

Our approach starts with an in-depth analysis of the observation from the behavior of the cross-attention layer during the diffusion process. 
Based on this analysis, we propose the Human-centric Prior layer with Human-centric Alignment loss to infuse human-centric prior knowledge. 
Subsequently, we detail the training strategy on both scale and step aware aspects.
Figure \ref{fig:framework} illustrates the procedure associated with the proposed HcP layer in the pre-trained latent diffusion model.

\vspace{3mm}
\tocless
\subsection{Analysis of Cross-Attention Layer}
\label{sec:insights}
\begin{figure}[h]
\centering
    \includegraphics[width=\linewidth]{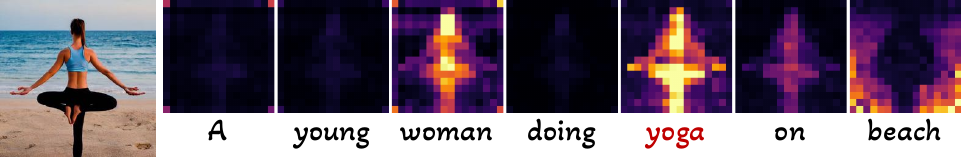}
    \caption{Average cross-attention maps across all timestamps of a text-conditioned diffusion process. These maps contain semantic relations with texts that affect the generated image, exemplified by the inaccurate duplication of legs in the generated human figure.}
    \label{fig:text}
\end{figure}

For the tHIG task, the aim is to generate a diverse set of images using a given text-to-image generation model driven by human-authored prompts.
However, there exist certain issues in the context of generating human images, such as structural inaccuracies and inconsistent body proportions. 
As demonstrated in \cite{hertz2022prompt}, the detailed structures in the generated images crucially depend on the interaction between the pixels and the text embedding at the cross-attention layers of U-Net. Consequently, we further examine the relationship between human body structures and each text token embedding in human image generation through the cross-attention process. For instance, in challenging cases like the prompt ``\textit{a young woman doing yoga on the beach}," we observe significant issues in rendering accurate human poses and proportions, as illustrated in Figure \ref{fig:text}.
Note that all observations and analysis are conducted on the publicly available Stable Diffusion v1-5 model \cite{rombach2022high}.

\begin{figure}[t]
\centering
    \includegraphics[width=\linewidth]{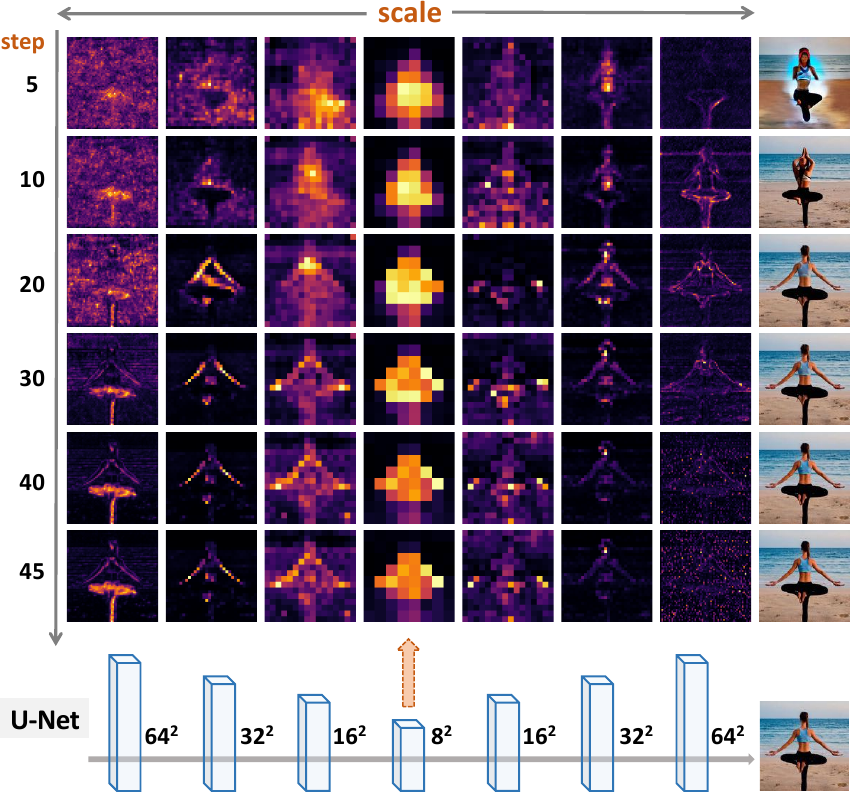}
    \caption{The cross-attention maps, as influenced by the fixed token 'yoga', are across various stages of the U-Net architecture at different inference timesteps. The vertical axis represents the inference timestep when using DDIM \cite{song2020denoising}, while the horizontal axis corresponds to the different scale stages within the U-Net framework. The right side displays generated images at each step.}
    \label{fig:ssinsights}
\end{figure}

We can see that the cross-attention map corresponding to ``\textit{woman}'' and ``\textit{yoga}'' closely reflects the human pose, and the map for ``\textit{beach}'' corresponds to background.
This strong correlation between attention maps and texts indicates that cross-attention layers, guided by specific text embeddings, play a pivotal role in shaping the semantic content of the image. This also implies that insufficient capabilities of cross-attention layers can affect the results of generated images.
Building on this observation, we conduct a comprehensive analysis as shown in Figure \ref{fig:ssinsights}, to identify and address the underlying causes of prevalent issues in tHIG.

\mypara{Step-wise Observation}.
The inference of the diffusion model is essentially a denoising process.
Given each step in the diffusion process incrementally refines the output image, it's essential to analyze the impact of early vs. later steps, especially in the context of human-centric image generation.
As illustrated in Figure \ref{fig:ssinsights}, the anatomical structure of the human subject becomes distinguishable in the very early steps. Later steps, while refining and enhancing the image, primarily focus on the optimization of finer details rather than significant structural alterations. This indicates the role of the initial steps is determining the overall structure and posture of the generated human figure, while later steps work on refining details to improve the final output.

\mypara{Scale-wise Observation}.
Based on our step-wise observations, we further investigate the role of resolution scale in synthesizing human images, particularly within the U-Net architecture of diffusion models. 
As illustrated in Figure \ref{fig:ssinsights}, we observe that as the resolution decreases (towards the middle of the U-Net architecture), mid-stage timesteps predominantly determine the structural aspects of the human figure. 
At the smaller resolution scale, located at the midpoint of the U-Net, all timesteps collectively influence the human structure, with early timesteps playing a more significant role.
Conversely, as the resolution increases again (moving towards the output layer), the early timesteps become key in defining the structure.
These observations underscore the complexity inherent in the cross-attention layers and the pivotal role of different scales and steps in the human image generation process.

% \vspace{3mm}
\tocless
\subsection{Human-centric Prior Layer}
\label{sec:hcp layer}

As we discussed in Section 
% \ref{sec:insights},
\textcolor{red}{3.1},
text embeddings related to humans and actions significantly influence the human structure in the generated image, which is particularly evident within the associated cross-attention maps. 
Therefore, we suggest that by enhancing the sensitivity of diffusion models to human-centric textual information during the denoising process, we can improve the structural accuracy and details in the generated images. To do this, we propose an additional learnable module, the Human-centric Prior (HcP) layer, to strengthen the interactions between the latent features and the human-centric textual within the cross-attention maps.
% Thus, to strengthen the interactions between the latent features and the human-centric textual information within these maps, we propose an additional learnable module, Human-centric Prior (HcP) layer. 
This module is integrated without altering the pre-existing expressive capacity of the cross-attention layers, whose parameters remain frozen during training. 
% Also, it is designed to specifically enhance the generation of human structure within the image synthesis process.

% Within the latent diffusion framework, the cross-attention layer of each U-Net block utilizes text token embeddings $\mathcal{C} = \{ \mathcal{C}_1, \mathcal{C}_2, \dots, \mathcal{C}_n \}$, $\mathcal{C} \in \mathbb{R}^{N \times D}$, to impose textual constraints on image generation, where $N$ and $D$ denote token length and embedding dimension, respectively.
% \textcolor{red}{Specifically, $\mathbf{Q}$ represents the latent representation, while both $\mathbf{K}$ and $\mathbf{V}$ are derived from the text-conditioned embeddings.}
Within the latent diffusion framework, the structure allows the cross-attention layer to effectively incorporate textual information into the image synthesis process.
Specifically, in this cross-attention mechanism, query $\mathbf{Q}$ represents the latent representation, capturing the spatial attributes at a specific resolution stage.
On the other hand, both key
$\mathbf{K}$ and value $\mathbf{V}$ are derived from the text-conditioned embeddings 
$\mathcal{C} = \{ \mathcal{C}_1, \mathcal{C}_2, \dots, \mathcal{C}_n \}$, $\mathcal{C} \in \mathbb{R}^{N \times D}$, where $N$ and $D$ denote text token length and embedding dimension. 
Subsequently, we introduce an additional ``Key" into the cross-attention mechanism, denoted as $\mathbf{K}_{h}$.
This key is also derived from the text embeddings $\mathcal{C}$ via the HcP layer which is composed of multiple MLP networks.
Then, $\mathbf{K}_{h}$ interacts with the Query, generating the human-centric attention map $\mathcal{M}_{h}$ as:
\begin{equation}
\mathcal{M}_{h} = \text{softmax} \left( \frac{\mathbf{Q} \mathbf{K}_{h}^T}{\sqrt{d}} \right),~\mathbf{K}_{h} = \phi(\mathcal{C}_h)~,
\end{equation}
where $\phi(\cdot)$ represents the transformation carried out by the HcP layer and $d$ indicates the latent projection dimension of the keys and queries.
Then the forward attention map of the cross-attention layer in the pre-trained denoising network is defined as the combination of the human-centric attention map $\mathcal{M}_{h}$ and the original attention map $\mathcal{M}$: 
\begin{equation}
    \hat{\mathcal{M}} = \gamma\mathcal{M} + (1-\gamma)\mathcal{M}_{h}~,
\label{equ:2}
\end{equation}
where $\gamma$ denotes the attention combination weight. 
Note that the HcP layer is a plug-and-play module that can be combined with any cross-attention layers.
This integration not only preserves the expressive power of the existing pre-trained U-Net, but also addresses the issues of human structure generation within the image synthesis process.
Subsequent subsections will describe the training process for the HcP layer to incorporate human-specific information.

\vspace{3mm}
\tocless
\subsection{Human-centric Alignment Loss}
\label{sec:hcp loss}

Acknowledging the diffusion model's deficiency in focusing on the details of human structure, we focus on enhancing human-specific information within the HcP layer.
Meanwhile, key pose images, effective in representing human body structures, are leveraged as essential sources of human-centric prior information.
Consequently, we have designed a novel loss function that aligns this human-centric prior information with the HcP layer, thereby addressing the structural challenges in human image generation.

Concretely, a pre-trained entity-relation network is first deployed to extract human-centric words from textual prompts. For instance, \textit{woman} and \textit{yoga} from the phrase ``\emph{A young woman doing yoga on beach}''.
Upon identifying human-centric terms, we only train corresponding indices within the attention map. This selective approach ensures the training focus of the human-centric attention map to the relevant regions.
% Meanwhile, the key pose image of human figures is adopted as a pivotal source of human-centric prior information in this work.
% We extract representative features $\mathbf{H}$, indicative of these poses, through a pre-trained encoder like ResNet50, providing a reference for human-centric characteristics.
We then utilize a pre-trained encoder, such as ResNet50, to extract features $\mathbf{H}$ from the corresponding key pose images that provide a reference for human-centric characteristics.
These features are aligned with the human-centric attention map $\mathcal{M}_{h}$, facilitated by a specially designed Human-centric Alignment Loss. This loss is computed using cosine distance, formulated as:
\begin{equation}
    \mathcal{L}_{hca}(\mathbf{H},\mathcal{M}_{h}) =  \frac{1}{|\mathcal{I}_h|} \sum_{i \in \mathcal{I}_h} [1 -\mathcal{D}(\mathbf{H},~\mathcal{M}_{h}[i])]~,
\end{equation}
where $\mathcal{D}(\cdot, \cdot)$ denotes the cosine similarity function and $|\mathcal{I}_h|$ is the count of human-centric word indices.
By minimizing the cosine distance in this manner, the human-centric attention map becomes more focused on human-centric prior information, as illustrated in the right part of Figure \ref{fig:framework}.
Notably, refinement is constrained to areas related to relevant tokens, with human-centric prior information directing the synthesis of human structures. 
% \textcolor{red}{write more here, e.g., analyze why we select cosine distance rather than MSE. Moreover, this design allows the inference process to operate without additional conditions.}

\vspace{3mm}
\tocless
\subsection{Scale \& Step Aware Learning}
\label{sec:sslearning}

% According to our observations from the cross-attention layer, we propose a targeted learning strategy designed to capitalize on both scale and step findings. This strategy involves a dual focus: one that addresses the scale-dependent impact of the U-Net architecture on the structural and detailed features of human figures, and another that adjusts the learning process to align with the step-wise refinement.
% Based on our detailed observations of the cross-attention layer's behavior in human image generation, we introduce a novel learning strategy that
% both the scale and step intricacies observed in the U-Net architecture.
% In the down-stage, as the resolution begins to decrease, the mid-stage timesteps emerge as key drivers in shaping the human figure's structural aspects. At the lowest resolution, occurring in the mid-stage, the influence is distributed across all timesteps, with a pronounced emphasis on the early timesteps. Conversely, in the up-stage, where the resolution starts to increase, the early timesteps quickly become pivotal in defining the human figure's structure.

\begin{figure}[t]
\centering
    \includegraphics[width=0.9\linewidth]{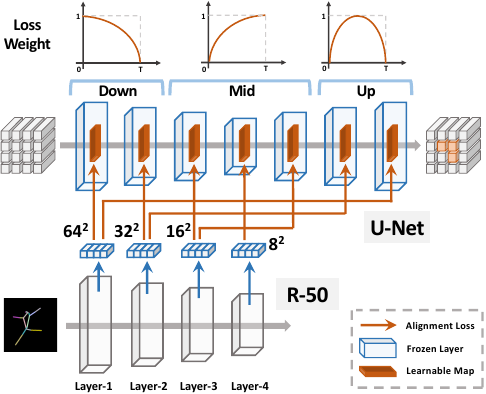}
    \caption{Alignment of layer-specific ResNet features with corresponding scale ([64$^2$,32$^2$,16$^2$,8$^2$]) human-centric attention maps in each cross-attention layer of the U-Net architecture for human-centric alignment loss}
    \label{fig:scale}
\end{figure}

Our detailed scale and step analysis in the inference phase (Section 
% \ref{sec:insights})
\textcolor{red}{3.1})
reveal a critical insight that the formation of human structure is closely linked to the resolution scale at different U-Net stages. 
Based on this observation, we introduce a learning strategy that addresses the unique scale and step characteristics observed in the U-Net architecture.
In this work, we first partition the U-Net of the Stable Diffusion v1-5 model into three distinct stages: \textit{down}, \textit{mid}, and \textit{up}. This partition reflects the different resolution scales within the U-Net, as shown in Figure \ref{fig:scale}.
% Specifically, the down and up stages are associated with higher resolution scales, while the mid-stage aligns with a lower resolution as shown in Figure \ref{fig:scale}.

% We employ a cosine function to dynamically modify loss weights for each stage.

% Concretely,  
% In constructing our model, we first segment the U-Net architecture in the SDv1-5 base model into three distinct stages: down, mid, and up. This segmentation reflects the diverse resolution scales within the U-Net. Specifically, the down and up stages are associated with higher resolution scales, while the mid stage aligns with a lower resolution. Our detailed analysis reveals a critical insight: the structural formation of the human figure is intricately tied to the resolution scale at different stages of the U-Net. In the down stage, as the resolution begins to decrease, the mid-stage timesteps emerge as key drivers in shaping the human figure's structural aspects. At the lowest resolution, occurring in the mid stage, the influence is distributed across all timesteps, with a pronounced emphasis on the early timesteps. Conversely, in the up stage, where the resolution starts to increase, the early timesteps quickly become pivotal in defining the human figure's structure. These insights underscore the importance of adjusting the learning focus according to the specific stage within the U-Net, prompting us to employ a cosine function to dynamically modify loss weights for each stage, as outlined in the following formula
In order to dynamically adjust the loss weights $\lambda$ at each stage of the U-Net, we utilize the \textbf{cosine function}, specifically adapted to the distinct characteristics of each scale.
The formula for this dynamic adjustment is expressed as:
\begin{equation}
    \lambda^l(t) = 
\begin{cases}
    \displaystyle\cos\left(\frac{t}{\mathbf{T}} \cdot \frac{\pi}{2}\right), & \text{if } l \in \text{down-scale} \\
    \displaystyle\cos\left(\frac{t - \mathbf{T}}{\mathbf{T}} \cdot \frac{\pi}{2}\right), & \text{if } l \in \text{mid-scale} \\
    % \displaystyle\cos\left(\frac{t - \mathbf{T}}{\mathbf{T}} \cdot \frac{\pi}{2}\right), & \text{if } l \in \text{down-scale} \\
    % \displaystyle\cos\left(\frac{t}{\mathbf{T}} \cdot \frac{\pi}{2}\right), & \text{if } l \in \text{mid-scale} \\
    \displaystyle\cos\left(\frac{2t - \mathbf{T}}{\mathbf{T}} \cdot \frac{\pi}{2}\right), & \text{if } l \in \text{up-scale}
\end{cases}
\end{equation}
where $l$ denotes the cross-attention layer number in U-Net.
For the down-scale stage, the loss weight follows a cosine function that varies in a straightforward manner with the progression of timestep $t$ relative to the maximum timestep $\mathbf{T}$.
This adjustment significantly impacts the human structural aspect at early timesteps.
For the mid-scale stage, where the resolution is lower, the loss weight is adjusted through a cosine function centered around the midpoint of the timesteps. This adjustment allows a higher emphasis on the later ones.
For the up-scale stage, as the resolution increases, the cosine function is designed to rapidly emphasize the middle timesteps, highlighting their importance in defining the human structure as the resolution scales up. 
% for a balanced influence of all timesteps

This strategy is designed to optimize the learning process by adapting to the specific requirements at different scales and steps, as revealed in our prior cross-attention map analysis.
It adjusts the learning focus, transitioning between structural definition and detailed texturing in accordance with the resolution scale.

\mypara{Overall Optimization}.
Meanwhile, the denoising loss is also incorporated into the training process to further ensure the quality of the synthesized images. 
Therefore, the overall optimization objective can be expressed as follows:
\begin{equation}
    \mathcal{L}_{ldm} = \mathbb{E}_{x,\epsilon \sim \mathcal{N}(0,1)}[\lVert \epsilon - \epsilon_{\theta}(z_t, t)\rVert_2^2]~,
\end{equation}
\begin{equation}
% \mathcal{L}^t = \lambda_{step}^t\sum_{l \in L}(\lambda_{scale}^l \cdot \mathcal{L}_{hca}^{l}) + \mathcal{L}_{ldm}
\mathcal{L}^t = \alpha\sum_{l \in L}(\lambda^l(t) \cdot \mathcal{L}_{hca}^{l}) + \mathcal{L}_{ldm}~,
\end{equation}
where $L$ denotes the number of U-Net layers and $\alpha$ denotes the human-centric alignment loss weight.
% In contrast to fine-tuning approaches such as LoRA \cite{hu2021lora}, our method preserves the original generative capabilities of the model without altering its expressive power. Instead, we focus on refining the human structure within the generated images to ensure a more reasonable representation. 
% Moreover, unlike methods that rely on additional conditions for image synthesis, like ControlNet \cite{zhang2023adding} and T2I-Adapter \cite{mou2023t2i}, our approach operates without the need for extra inputs, thereby maintaining diversity in the generative process. 
In contrast to other approaches, our method preserves the original generative capabilities of the model without altering its expressive power and focuses on refining the human structure within the generated images to ensure a more reasonable representation. 
Meanwhile, it operates without extra inputs, thereby maintaining diversity in the generative process.

% \newpage
% \begin{figure*}[h]
% \centering
%     \includegraphics[width=\linewidth]{figs/mainresults.pdf}
%     % \fbox{\rule{0pt}{2in} \rule{.9\linewidth}{0pt}}
%     \caption{Qualitative Evaluation.}
%     \label{fig:qualitative}
% \end{figure*}
\vspace{3mm}
\tocless
\section{Experiments}
\label{sec:experiments}
We validate the proposed HcP layer for HIG in various scenarios and introduce the experimental setup in Section
% \ref{subsec:setup},
\textcolor{red}{4.1},
presents the main results in Section
% ~\ref{subsec:main},
\textcolor{red}{4.2},
and detailed ablations and discussions in Section
% ~\ref{subsec:ablation} 
\textcolor{red}{4.3}
and
% ~\ref{subsec:discussion}.
\textcolor{red}{4.4}.
Please see the Appendix for additional results and analyses.

\vspace{3mm}
\tocless
\subsection{Setup}
\label{subsec:setup}
\noindent\textbf{Datasets.}
(1) \emph{Human-Art} \cite{ju2023human} contains 50k images in 20 natural and artificial scenarios with clean annotation of pose and text, which provide precise poses and multi-scenario for both training and quantitative evaluation. 
(2) \emph{Laion-Human} \cite{ju2023human} contains 1M image-text pairs collected from LAION-5B \cite{schuhmann2022laion} filtered by the rules of high image quality and high human estimation confidence scores.

\mypara{Evaluation Metrics.}
To comprehensively illustrate the effectiveness of our proposed method, we adopt three different types of metrics:
(1) \emph{Image Quality}: Frechet Inception Distance (FID) \cite{heusel2017gans} and Kernel Inception Distance (KID) \cite{binkowski2018demystifying} to measure the quality of the syntheses. 
(2) \emph{Text-image Consistency}: CLIP-Score \cite{radford2021learning} to evaluate text-image consistency between the generated images and corresponding text prompts. 
(3) \emph{Human Evaluation}: This further evaluates the anatomy quality and examines the consistency between the text-image pairs using human's subjective perceptions.

\mypara{Baselines.}
We compare HcP layer to the following methods. 
(1) \textit{Stable Diffusion} (SD) v1-5 \cite{rombach2022high} without any modification.
(2) \textit{Low-rank Adaptation} (LoRA) \cite{hu2021lora} fine-tuned with SD model on both Human-Art training set and Laion-Human set.
Additionally, we also compare with \textit{ControlNet} \cite{zhang2023adding} using the \textit{OpenPose} condition, and \textit{SDXL-base} \cite{podell2023sdxl}.

\mypara{Implementation Details.}
The total trainable parameters are from the proposed HcP layer which consists of three 1024-dimensional MLP blocks.
We choose key pose image as human-centric prior information and use pre-trained ResNet-50 \cite{he2016deep} as the human-centric prior information extractor.
To align the scale of each layer's features of ResNet-50 with those from the cross-attention layer in U-Net, the input pose images are resized to 256 $\times$ 256.
We select the top eight features with the highest variance across channels from the last four stages of ResNet50. These are leveraged as the multi-heads for the cross-attention layer of U-Net, with the head number set to 8.
During training, we use the AdamW optimizer \cite{loshchilov2018decoupled} with a fixed learning rate of 0.0001 and weight decay of 0.01, and we set $\gamma = 0.9$ and $\alpha = 0.1$ for loss control.
In the inference stage, we adopt DDIM sampler \cite{song2020denoising} with 50 steps and set the guidance scale to 7.5.
All experiments are performed on 8 $\times$ Nvidia Tesla A100 GPUs.
More implementation details in Appendix \ref{app: setting}.

\begin{figure}[t]
\centering
    \includegraphics[width=\linewidth]{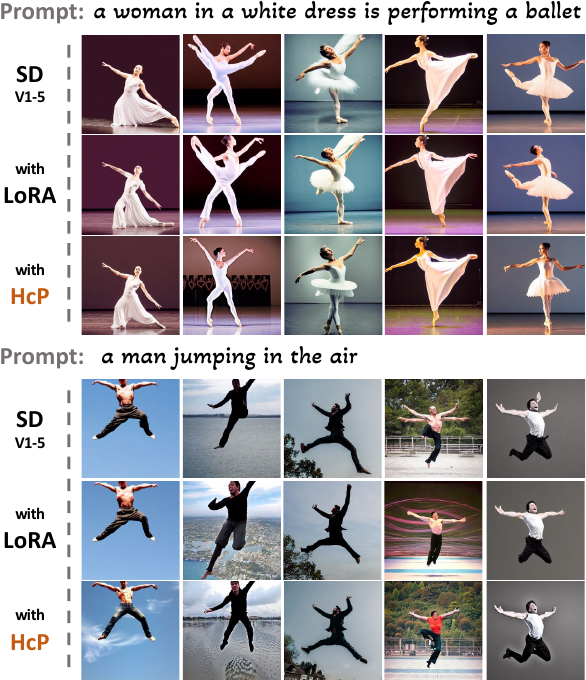}
    \caption{\textbf{Qualitative comparison with baseline methods on two example prompts}. 
    We leverage the pre-trained SD v1-5 model for both ``with LoRA" and ``with HcP" models while keeping it frozen. More examples across domains are included in the Appendix \ref{app: qualitative}.}
    \label{fig:qualitative}
\end{figure}

\vspace{3mm}
\tocless
\subsection{Main Results}
\label{subsec:main}
We validate the superiority of the HcP layer by combining with pre-trained SD and making comparisons with vanilla SD, and SD enhanced with LoRA, from both qualitative and quantitative perspectives.

\mypara{Qualitative Evaluation}.
As shown in Figure \ref{fig:qualitative}, for simpler actions like ``\textit{jumping}", the pre-trained SD enhanced with LoRA shows improved quality in human image generation, but its effectiveness diminishes with more complex actions such as ``\textit{ballet}".
Furthermore, LoRA somehow alters the depiction of the original diffusion model, especially for background content, indicating that it enhances the generation of human structures while simultaneously affecting the model's intrinsic capability to represent scenes.
In contrast, our proposed method with the HcP layer shows consistently accurate human structure generation across a variety of actions, both simple and complex. 
Notably, our method retains the original expressive power of the pre-trained SD more effectively, maintaining both the background content and human structure more closely aligned with the original model, reflecting a more focused enhancement.
This evaluation demonstrates the effectiveness of the HcP layer in addressing human image structure issues without significantly altering the model's overall image synthesis capabilities.

\begin{table}[t]
    \centering
        \caption{\textbf{FID, KID, and CLIP-Score results on Human-Art validation datasets.} $\downarrow$ indicates that lower FID and KID are better, reflecting higher image quality; $\uparrow$ denotes higher CLIP-Score indicating better alignment with textual descriptions.
        }\vspace{-2mm}
    \scalebox{0.8}{
    \begin{tabular}{l c c c c}
      \multirow{2}{*}{Method} &
        \multicolumn{2}{c}{Quality}  &
        \multicolumn{1}{c}{Consistency}      \\
        \multicolumn{1}{c}{} &
        \multicolumn{1}{c}{FID $\downarrow$}   &
        \multicolumn{1}{c}{KID$_{\times 1k}$ $\downarrow$}   &
        \multicolumn{1}{c}{CLIP-Score $\uparrow$}   \\
      \midrule
        SD &  33.31 & 9.38 & 31.85  \\
        + LoRA &   29.22 & 5.83 & 31.91 \\
        \midrule
        + HcP &   \textbf{28.71} & \textbf{5.62} & \textbf{32.72} \\
    \end{tabular}
    }
    \label{tab:quantitative}\vspace{-2mm}
\end{table}

\begin{table}[t]
    \centering
        \caption{\textbf{Human evaluation on the real-human category of Human-Art dataset.} Participants were asked to rate every pair by using a 5-point Likert scale (1 = poor, 5 = excellent), considering \emph{anatomy quality} (AQ) and \emph{text-image alignment} (TIA).}\vspace{-2mm}
    \scalebox{0.73}{
    \begin{tabular}{l c c c c c c c c c c}
      \multirow{2}{*}{Method} &
        \multicolumn{2}{c}{Acrobatics} &
        \multicolumn{2}{c}{Cosplay}  &
        \multicolumn{2}{c}{Dance}  &    
        \multicolumn{2}{c}{Drama} &
        \multicolumn{2}{c}{Movie} \\
        \multicolumn{1}{c}{}   &
        \multicolumn{1}{c}{AQ}   &
        \multicolumn{1}{c}{TIA}   &
        \multicolumn{1}{c}{AQ}   &
        \multicolumn{1}{c}{TIA}   &
        \multicolumn{1}{c}{AQ}   &
        \multicolumn{1}{c}{TIA}   &
        \multicolumn{1}{c}{AQ}   &
        \multicolumn{1}{c}{TIA}   &
        \multicolumn{1}{c}{AQ}   &
        \multicolumn{1}{c}{TIA}   \\
      \midrule
        SD & 1.6 & 2.2 & 3.5 & 4.1 & 2.0 & 2.5 & 2.0 & 1.8 & 3.0 & 3.4\\
        + LoRA & 1.8 & 2.2 & 3.6 & 4.1 & 2.1 & 2.5 & 2.0 & 2.5 & 3.0 & 3.5  \\
        \midrule
        + HcP & \textbf{2.7} & \textbf{3.5} & \textbf{3.8} & \textbf{4.3} & \textbf{3.5} & \textbf{4.0} & \textbf{3.2} & \textbf{2.6} & \textbf{3.1} & \textbf{3.6} \\
    \end{tabular}
    }
    \label{tab:huamneval} \vspace{-2mm}
\end{table}

\mypara{Quantitative Evaluation}.
According to the results in Table \ref{tab:quantitative}, the image quality metrics reveal that our HcP method does not compromise the original generation quality of the SD v1-5 model. 
Furthermore, our approach achieves a more significant increase in CLIP-Score compared with LoRA fine-tuning. 
This improvement underscores the efficacy of the HcP layer in refining human structure generation, ensuring a more accurate depiction of human poses and proportions in alignment with the textual descriptions.

\mypara{Human Evaluation}.
To further verify the efficacy of HcP, we invited participants to evaluate our prompt-generated image pairs under the guidelines of multimedia subjective testing~\cite{ITU}.
To be specific, we use different methods to generate 200 images for different domains in the Human-Art dataset.
The results presented in Table \ref{tab:huamneval} demonstrate a significant improvement in generating human figures for complex domains (`acrobatics', `dance', and `drama') using our method, compared to both SD and LoRA. Additionally, our approach yields comparable results to these methods in simpler domains (`cosplay' and `movie'). These findings further validate the effectiveness of our proposed method in improving the capability of the diffusion model to produce a more accurate human structure and meanwhile retaining the original expressive power of the pre-trained diffusion model.
More details can be seen in Appendix \ref{app: human}.

\vspace{3mm}
\tocless
\subsection{Ablation Study}
\label{subsec:ablation}

\begin{figure}[t]
\centering
    \includegraphics[width=\linewidth]{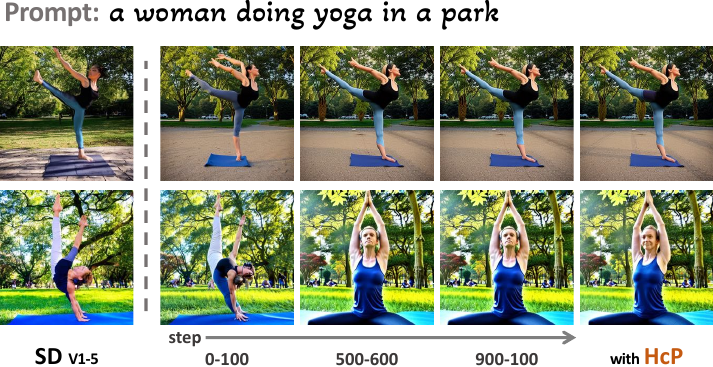}
    \caption{
    \textbf{Ablation on different timestamp stages}. 
    The middle three images are the outcomes of training the model in three distinct phases (0-100, 500-600, and 900-1000 timesteps) without 
    the cosine function for scale adjustments.}
    \label{fig:abstep}
\end{figure}

\begin{figure}[t]
\centering
    \includegraphics[width=\linewidth]{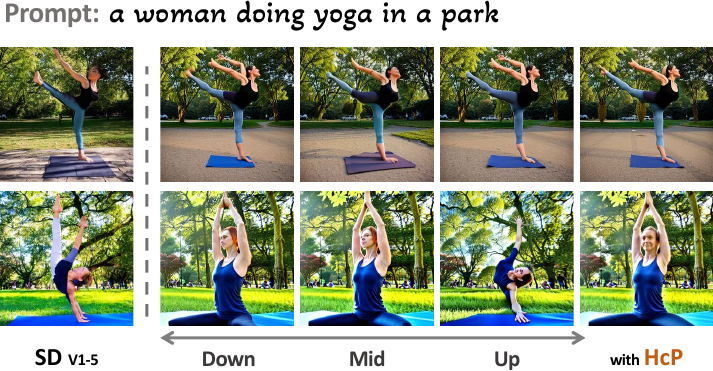}
    \caption{
    \textbf{Ablation on cosine function in different scale stages}. 
    The middle three images are the outcomes of training the model at the down, mid, and up scales without cosine function adjustments.}
    \label{fig:abscale}
\end{figure}

\mypara{Different Timestamp Stages}.
During the training phase with the DDPM, which involves a maximum of 1000 timesteps,
we selectively apply the human-centric alignment loss in different time segments: \textit{early}, \textit{middle}, and \textit{late} phases, as shown in Figure \ref{fig:abstep}.
When the human-centric alignment loss is introduced during the early timesteps, the influence on human image generation is comparatively minimal.
Essentially, applying the alignment loss too early fails to fully leverage the human-centric prior information.
Conversely, when applied during the mid or late timesteps, the human-centric alignment loss affects the generation of human structures. 
It leads to the creation of more accurate human images through efficiently utilizing human-centric prior information. 
This finding aligns with our inference stage observation in Section
% ~\ref{sec:insights},
\textcolor{red}{3.1},
which the initial steps are crucial in establishing the overall structure and posture of the generated human image, while later steps work on refining details to improve the quality of the final output.

\mypara{Scale-Aware Training}. 
In this validation, we separately excluded the cosine function adjustment at the down-scale, mid-scale, or up-scale stages of the U-Net, as results shown in Figure \ref{fig:abscale}.
As illustrated, the absence of the cosine function adjustment in the mid-scale leads to outcomes nearly unchanged from the final images, though with certain limitations.
This corroborates our observation that at the smaller resolution scale, all timesteps collectively contribute to shaping the human structure.
Significant deviations in results are observed when the cosine function adjustment is not applied in either the up or down scales, especially in the up-scale, which reinforces our observation regarding the distinct influence of different scale stages.
Meanwhile, these further validate the appropriateness of applying cosine function adjustments at each scale in the U-Net architecture.

\vspace{3mm}
\tocless
\subsection{Discussion}
\label{subsec:discussion}

\begin{figure}[t]
\centering
    \includegraphics[width=\linewidth]{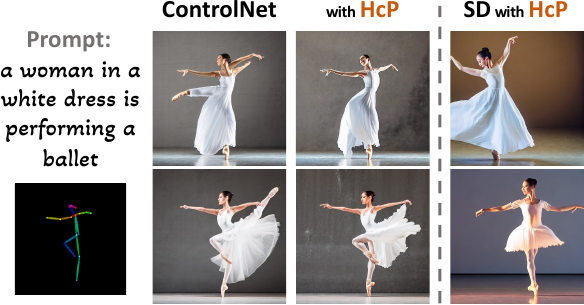}
    \caption{
    \textbf{Comparisons and compatibility with the controllable HIG application}.
    By plugging the HcP layer trained on pre-trained SD into ControlNet~\cite{zhang2023adding}, our method can further boost both the quality and consistency compared with the original ControlNet-OpenPose model. 
    }
    \label{fig:controlnet}
\end{figure}
\begin{figure}[t]
\centering
    \includegraphics[width=\linewidth]{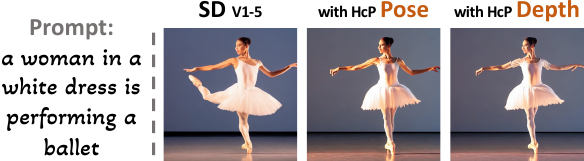}
    \caption{\textbf{Comparisons by using different sources of human-centric prior information}. The middle and right images utilize pose and depth images as the human-centric prior information respectively. More results can be seen in Appendix \ref{app: depth}.
    }
    \label{fig:depth}
    \vspace{-1mm}
\end{figure}

\mypara{Controllable HIG}.
Considering the adaptable design of our proposed HcP layer as a plug-and-play approach, it can also be extended to Controllable HIG applications. According to Figure \ref{fig:controlnet}, despite having a defined pose, ControlNet still encounters challenges in accurately generating human structures. 
Interestingly, by simply plugging the proposed HcP layer, which is fine-tuned only on the SD instead of the ControlNet model, into the ControlNet, human images with more precise structure and posture are obtained.
Moreover, even utilizing only the pre-trained SD model with the HcP layer without relying on any additional condition in the inference phase, our method can acquire comparable results and ensure diverse generations based on only textual inputs.
More comparisons can be seen in Appendix \ref{app: controlnet}.

\mypara{Human-centric Prior Information}.
In Figure~\ref{fig:depth}, we utilize depth images as an alternative source of human-centric prior information.
The results demonstrate that depth images are also effective in correcting inaccuracies in human image generation.
While depth prior can enhance the detailing of the generated images, they tend to slightly alter the details of the original human image, such as the textures of clothing, compared to pose images.
In future work, we plan to investigate how to use 
multiple types of human-centric prior information to optimize the balance between detail enhancement and structural accuracy for generated images.

\begin{figure}[t]
\centering
    \includegraphics[width=\linewidth]{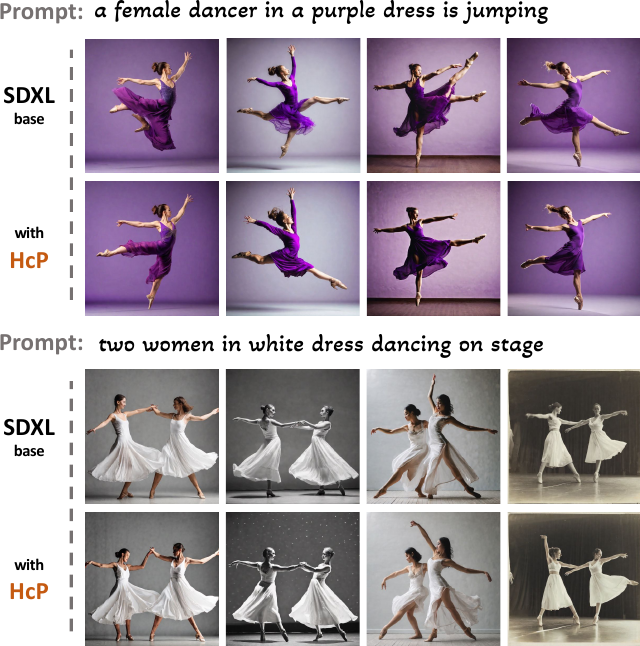}
    \caption{\textbf{Results on larger diffusion model using HcP layer.} We leverage the pre-trained SDXL-base model for the ``with HcP" model while keeping it frozen.
    More examples of different scenarios are included in the Appendix \ref{app: qualitative}.}
    \label{fig:sdxl}
    \vspace{-2mm}
\end{figure}

\mypara{Large Diffusion Model}.
To assess the effectiveness of our method on larger vision models, we evaluated it on SDXL-base, as shown in Figure \ref{fig:sdxl}.
The results demonstrate that, while SDXL generally produces human images with better structure and detail compared to SD v1-5, it still exhibits some issues. For example, the proportions of the legs are not harmonious in the first image, and extra legs are in other figures.
Notably, our method not only capably addresses these issues on the larger model but also enhances the overall fidelity and precision of the generated images.

\vspace{3mm}
\tocless
\section{Conclusion}
\label{sec:conclusions}
In this work, we propose a simple yet effective method of using human-centric priors (HcP), \eg pose or depth maps, to improve human image generation in existing text-to-image models.
The proposed HcP layer effectively uses information about humans during the fine-tuning process without needing extra input when generating images from text.
Extensive experiments demonstrate that the HcP layer not only fixes structural inaccuracies in human structure generation but also preserves the original aesthetic qualities and details.
Future work will explore the integration of multiple types of human-centric priors to further advance human image and video generation.

% refer
{
    \small
    \bibliographystyle{ieeenat_fullname}
    \bibliography{main}

\begin{thebibliography}{55}
\providecommand{\natexlab}[1]{#1}
\providecommand{\url}[1]{\texttt{#1}}
\expandafter\ifx\csname urlstyle\endcsname\relax
  \providecommand{\doi}[1]{doi: #1}\else
  \providecommand{\doi}{doi: \begingroup \urlstyle{rm}\Url}\fi

\bibitem[Avrahami et~al.(2022)Avrahami, Lischinski, and Fried]{avrahami2022blended}
Omri Avrahami, Dani Lischinski, and Ohad Fried.
\newblock Blended diffusion for text-driven editing of natural images.
\newblock In \emph{CVPR}, pages 18208--18218, 2022.

\bibitem[Bi{\'n}kowski et~al.(2018)Bi{\'n}kowski, Sutherland, Arbel, and Gretton]{binkowski2018demystifying}
Miko{\l}aj Bi{\'n}kowski, Danica~J Sutherland, Michael Arbel, and Arthur Gretton.
\newblock Demystifying mmd gans.
\newblock In \emph{ICLR}, 2018.

\bibitem[Chang et~al.(2023)Chang, Zhang, Barber, Maschinot, Lezama, Jiang, Yang, Murphy, Freeman, Rubinstein, et~al.]{chang2023muse}
Huiwen Chang, Han Zhang, Jarred Barber, AJ Maschinot, Jose Lezama, Lu Jiang, Ming-Hsuan Yang, Kevin Murphy, William~T Freeman, Michael Rubinstein, et~al.
\newblock Muse: Text-to-image generation via masked generative transformers.
\newblock \emph{arXiv preprint arXiv:2301.00704}, 2023.

\bibitem[Chen et~al.(2023)Chen, Xu, Jia, Luo, Wang, Wang, Jin, and Sun]{chen2023beyond}
Weihua Chen, Xianzhe Xu, Jian Jia, Hao Luo, Yaohua Wang, Fan Wang, Rong Jin, and Xiuyu Sun.
\newblock Beyond appearance: a semantic controllable self-supervised learning framework for human-centric visual tasks.
\newblock In \emph{CVPR}, pages 15050--15061, 2023.

\bibitem[Croitoru et~al.(2023)Croitoru, Hondru, Ionescu, and Shah]{croitoru2023diffusion}
Florinel-Alin Croitoru, Vlad Hondru, Radu~Tudor Ionescu, and Mubarak Shah.
\newblock Diffusion models in vision: A survey.
\newblock \emph{IEEE Transactions on Pattern Analysis and Machine Intelligence}, 2023.

\bibitem[Dhariwal and Nichol(2021)]{dhariwal2021diffusion}
Prafulla Dhariwal and Alexander Nichol.
\newblock Diffusion models beat gans on image synthesis.
\newblock \emph{NeurIPS}, 34:\penalty0 8780--8794, 2021.

\bibitem[Ding et~al.(2022)Ding, Zheng, Hong, and Tang]{ding2022cogview2}
Ming Ding, Wendi Zheng, Wenyi Hong, and Jie Tang.
\newblock Cogview2: Faster and better text-to-image generation via hierarchical transformers.
\newblock \emph{NeurIPS}, 35:\penalty0 16890--16902, 2022.

\bibitem[document Rec. ITU-R(2007)]{ITU}
document Rec. ITU-R.
\newblock Methodology for the subjective assessment of video quality in multimedia applications.
\newblock \emph{BT.1788}, pages 1--13, 2007.

\bibitem[Esser et~al.(2021)Esser, Rombach, and Ommer]{esser2021taming}
Patrick Esser, Robin Rombach, and Bjorn Ommer.
\newblock Taming transformers for high-resolution image synthesis.
\newblock In \emph{CVPR}, pages 12873--12883, 2021.

\bibitem[Feng et~al.(2022)Feng, He, Fu, Jampani, Akula, Narayana, Basu, Wang, and Wang]{feng2022training}
Weixi Feng, Xuehai He, Tsu-Jui Fu, Varun Jampani, Arjun~Reddy Akula, Pradyumna Narayana, Sugato Basu, Xin~Eric Wang, and William~Yang Wang.
\newblock Training-free structured diffusion guidance for compositional text-to-image synthesis.
\newblock In \emph{ICLR}, 2022.

\bibitem[Han et~al.(2022)Han, Yu, Zhu, Zhang, Song, and Xiang]{han2022fashionvil}
Xiao Han, Licheng Yu, Xiatian Zhu, Li Zhang, Yi-Zhe Song, and Tao Xiang.
\newblock Fashionvil: Fashion-focused vision-and-language representation learning.
\newblock In \emph{ECCV}, pages 634--651. Springer, 2022.

\bibitem[Han et~al.(2023)Han, Zhu, Yu, Zhang, Song, and Xiang]{han2023fame}
Xiao Han, Xiatian Zhu, Licheng Yu, Li Zhang, Yi-Zhe Song, and Tao Xiang.
\newblock Fame-vil: Multi-tasking vision-language model for heterogeneous fashion tasks.
\newblock In \emph{CVPR}, pages 2669--2680, 2023.

\bibitem[He et~al.(2016)He, Zhang, Ren, and Sun]{he2016deep}
Kaiming He, Xiangyu Zhang, Shaoqing Ren, and Jian Sun.
\newblock Deep residual learning for image recognition.
\newblock In \emph{CVPR}, pages 770--778, 2016.

\bibitem[Hertz et~al.(2022)Hertz, Mokady, Tenenbaum, Aberman, Pritch, and Cohen-or]{hertz2022prompt}
Amir Hertz, Ron Mokady, Jay Tenenbaum, Kfir Aberman, Yael Pritch, and Daniel Cohen-or.
\newblock Prompt-to-prompt image editing with cross-attention control.
\newblock In \emph{ICLR}, 2022.

\bibitem[Heusel et~al.(2017)Heusel, Ramsauer, Unterthiner, Nessler, and Hochreiter]{heusel2017gans}
Martin Heusel, Hubert Ramsauer, Thomas Unterthiner, Bernhard Nessler, and Sepp Hochreiter.
\newblock Gans trained by a two time-scale update rule converge to a local nash equilibrium.
\newblock \emph{NeurIPS}, 30, 2017.

\bibitem[Ho et~al.(2020)Ho, Jain, and Abbeel]{ho2020denoising}
Jonathan Ho, Ajay Jain, and Pieter Abbeel.
\newblock Denoising diffusion probabilistic models.
\newblock \emph{NeurIPS}, 33:\penalty0 6840--6851, 2020.

\bibitem[Hu et~al.(2021)Hu, Wallis, Allen-Zhu, Li, Wang, Wang, Chen, et~al.]{hu2021lora}
Edward~J Hu, Phillip Wallis, Zeyuan Allen-Zhu, Yuanzhi Li, Shean Wang, Lu Wang, Weizhu Chen, et~al.
\newblock Lora: Low-rank adaptation of large language models.
\newblock In \emph{ICLR}, 2021.

\bibitem[Ju et~al.(2023{\natexlab{a}})Ju, Zeng, Wang, Xu, and Zhang]{ju2023human}
Xuan Ju, Ailing Zeng, Jianan Wang, Qiang Xu, and Lei Zhang.
\newblock Human-art: A versatile human-centric dataset bridging natural and artificial scenes.
\newblock In \emph{CVPR}, pages 618--629, 2023{\natexlab{a}}.

\bibitem[Ju et~al.(2023{\natexlab{b}})Ju, Zeng, Zhao, Wang, Zhang, and Xu]{ju2023humansd}
Xuan Ju, Ailing Zeng, Chenchen Zhao, Jianan Wang, Lei Zhang, and Qiang Xu.
\newblock Humansd: A native skeleton-guided diffusion model for human image generation.
\newblock \emph{arXiv preprint arXiv:2304.04269}, 2023{\natexlab{b}}.

\bibitem[Kim et~al.(2022)Kim, Kwon, and Ye]{kim2022diffusionclip}
Gwanghyun Kim, Taesung Kwon, and Jong~Chul Ye.
\newblock Diffusionclip: Text-guided diffusion models for robust image manipulation.
\newblock In \emph{CVPR}, pages 2426--2435, 2022.

\bibitem[Kingma et~al.(2021)Kingma, Salimans, Poole, and Ho]{kingma2021variational}
Diederik Kingma, Tim Salimans, Ben Poole, and Jonathan Ho.
\newblock Variational diffusion models.
\newblock \emph{NeurIPS}, 34:\penalty0 21696--21707, 2021.

\bibitem[Liu et~al.(2023{\natexlab{a}})Liu, Ren, Siarohin, Skorokhodov, Li, Lin, Liu, Liu, and Tulyakov]{liu2023hyperhuman}
Xian Liu, Jian Ren, Aliaksandr Siarohin, Ivan Skorokhodov, Yanyu Li, Dahua Lin, Xihui Liu, Ziwei Liu, and Sergey Tulyakov.
\newblock Hyperhuman: Hyper-realistic human generation with latent structural diffusion.
\newblock \emph{arXiv preprint arXiv:2310.08579}, 2023{\natexlab{a}}.

\bibitem[Liu et~al.(2023{\natexlab{b}})Liu, Feng, Zhu, Zhang, Zheng, Liu, Zhao, Zhou, and Cao]{liu2023cones}
Zhiheng Liu, Ruili Feng, Kai Zhu, Yifei Zhang, Kecheng Zheng, Yu Liu, Deli Zhao, Jingren Zhou, and Yang Cao.
\newblock Cones: Concept neurons in diffusion models for customized generation.
\newblock \emph{arXiv preprint arXiv:2303.05125}, 2023{\natexlab{b}}.

\bibitem[Loshchilov and Hutter(2018)]{loshchilov2018decoupled}
Ilya Loshchilov and Frank Hutter.
\newblock Decoupled weight decay regularization.
\newblock In \emph{ICLR}, 2018.

\bibitem[Mansimov et~al.(2015)Mansimov, Parisotto, Ba, and Salakhutdinov]{mansimov2015generating}
Elman Mansimov, Emilio Parisotto, Jimmy~Lei Ba, and Ruslan Salakhutdinov.
\newblock Generating images from captions with attention.
\newblock In \emph{ICLR}, 2015.

\bibitem[Mou et~al.(2023)Mou, Wang, Xie, Zhang, Qi, Shan, and Qie]{mou2023t2i}
Chong Mou, Xintao Wang, Liangbin Xie, Jian Zhang, Zhongang Qi, Ying Shan, and Xiaohu Qie.
\newblock T2i-adapter: Learning adapters to dig out more controllable ability for text-to-image diffusion models.
\newblock \emph{arXiv preprint arXiv:2302.08453}, 2023.

\bibitem[Pan et~al.(2022)Pan, Qin, Li, Xue, and Chen]{pan2022synthesizing}
Xichen Pan, Pengda Qin, Yuhong Li, Hui Xue, and Wenhu Chen.
\newblock Synthesizing coherent story with auto-regressive latent diffusion models.
\newblock \emph{arXiv preprint arXiv:2211.10950}, 2022.

\bibitem[Podell et~al.(2023)Podell, English, Lacey, Blattmann, Dockhorn, M{\"u}ller, Penna, and Rombach]{podell2023sdxl}
Dustin Podell, Zion English, Kyle Lacey, Andreas Blattmann, Tim Dockhorn, Jonas M{\"u}ller, Joe Penna, and Robin Rombach.
\newblock Sdxl: improving latent diffusion models for high-resolution image synthesis.
\newblock \emph{arXiv preprint arXiv:2307.01952}, 2023.

\bibitem[Qiao et~al.(2019)Qiao, Zhang, Xu, and Tao]{qiao2019mirrorgan}
Tingting Qiao, Jing Zhang, Duanqing Xu, and Dacheng Tao.
\newblock Mirrorgan: Learning text-to-image generation by redescription.
\newblock In \emph{CVPR}, pages 1505--1514, 2019.

\bibitem[Radford et~al.(2021)Radford, Kim, Hallacy, Ramesh, Goh, Agarwal, Sastry, Askell, Mishkin, Clark, et~al.]{radford2021learning}
Alec Radford, Jong~Wook Kim, Chris Hallacy, Aditya Ramesh, Gabriel Goh, Sandhini Agarwal, Girish Sastry, Amanda Askell, Pamela Mishkin, Jack Clark, et~al.
\newblock Learning transferable visual models from natural language supervision.
\newblock In \emph{ICML}, pages 8748--8763. PMLR, 2021.

\bibitem[Ramesh et~al.(2021)Ramesh, Pavlov, Goh, Gray, Voss, Radford, Chen, and Sutskever]{ramesh2021zero}
Aditya Ramesh, Mikhail Pavlov, Gabriel Goh, Scott Gray, Chelsea Voss, Alec Radford, Mark Chen, and Ilya Sutskever.
\newblock Zero-shot text-to-image generation.
\newblock In \emph{ICML}, pages 8821--8831. PMLR, 2021.

\bibitem[Reed et~al.(2016)Reed, Akata, Yan, Logeswaran, Schiele, and Lee]{reed2016generative}
Scott Reed, Zeynep Akata, Xinchen Yan, Lajanugen Logeswaran, Bernt Schiele, and Honglak Lee.
\newblock Generative adversarial text to image synthesis.
\newblock In \emph{ICML}, pages 1060--1069. PMLR, 2016.

\bibitem[Ren et~al.(2020)Ren, Yu, Chen, Li, and Li]{ren2020deep}
Yurui Ren, Xiaoming Yu, Junming Chen, Thomas~H Li, and Ge Li.
\newblock Deep image spatial transformation for person image generation.
\newblock In \emph{CVPR}, pages 7690--7699, 2020.

\bibitem[Ren et~al.(2022)Ren, Fan, Li, Liu, and Li]{ren2022neural}
Yurui Ren, Xiaoqing Fan, Ge Li, Shan Liu, and Thomas~H Li.
\newblock Neural texture extraction and distribution for controllable person image synthesis.
\newblock In \emph{CVPR}, pages 13535--13544, 2022.

\bibitem[Rombach et~al.(2022)Rombach, Blattmann, Lorenz, Esser, and Ommer]{rombach2022high}
Robin Rombach, Andreas Blattmann, Dominik Lorenz, Patrick Esser, and Bj{\"o}rn Ommer.
\newblock High-resolution image synthesis with latent diffusion models.
\newblock In \emph{CVPR}, pages 10684--10695, 2022.

\bibitem[Saharia et~al.(2022)Saharia, Chan, Saxena, Li, Whang, Denton, Ghasemipour, Gontijo~Lopes, Karagol~Ayan, Salimans, et~al.]{saharia2022photorealistic}
Chitwan Saharia, William Chan, Saurabh Saxena, Lala Li, Jay Whang, Emily~L Denton, Kamyar Ghasemipour, Raphael Gontijo~Lopes, Burcu Karagol~Ayan, Tim Salimans, et~al.
\newblock Photorealistic text-to-image diffusion models with deep language understanding.
\newblock \emph{NeurIPS}, 35:\penalty0 36479--36494, 2022.

\bibitem[Schuhmann et~al.(2022)Schuhmann, Beaumont, Vencu, Gordon, Wightman, Cherti, Coombes, Katta, Mullis, Wortsman, et~al.]{schuhmann2022laion}
Christoph Schuhmann, Romain Beaumont, Richard Vencu, Cade Gordon, Ross Wightman, Mehdi Cherti, Theo Coombes, Aarush Katta, Clayton Mullis, Mitchell Wortsman, et~al.
\newblock Laion-5b: An open large-scale dataset for training next generation image-text models.
\newblock \emph{NeurIPS}, 35:\penalty0 25278--25294, 2022.

\bibitem[Song et~al.(2020)Song, Meng, and Ermon]{song2020denoising}
Jiaming Song, Chenlin Meng, and Stefano Ermon.
\newblock Denoising diffusion implicit models.
\newblock \emph{arXiv preprint arXiv:2010.02502}, 2020.

\bibitem[Tao et~al.(2022)Tao, Tang, Wu, Jing, Bao, and Xu]{tao2022df}
Ming Tao, Hao Tang, Fei Wu, Xiao-Yuan Jing, Bing-Kun Bao, and Changsheng Xu.
\newblock Df-gan: A simple and effective baseline for text-to-image synthesis.
\newblock In \emph{CVPR}, pages 16515--16525, 2022.

\bibitem[von Platen et~al.(2022)von Platen, Patil, Lozhkov, Cuenca, Lambert, Rasul, Davaadorj, and Wolf]{von2022diffusers}
Patrick von Platen, Suraj Patil, Anton Lozhkov, Pedro Cuenca, Nathan Lambert, Kashif Rasul, Mishig Davaadorj, and Thomas Wolf.
\newblock Diffusers: State-of-the-art diffusion models, 2022.

\bibitem[Xu et~al.(2018)Xu, Zhang, Huang, Zhang, Gan, Huang, and He]{xu2018attngan}
Tao Xu, Pengchuan Zhang, Qiuyuan Huang, Han Zhang, Zhe Gan, Xiaolei Huang, and Xiaodong He.
\newblock Attngan: Fine-grained text to image generation with attentional generative adversarial networks.
\newblock In \emph{CVPR}, pages 1316--1324, 2018.

\bibitem[Xu et~al.(2024)Xu, Li, Tao, Shen, Cheng, Li, Xu, Tao, and Zhou]{xu2024survey}
Xiaohan Xu, Ming Li, Chongyang Tao, Tao Shen, Reynold Cheng, Jinyang Li, Can Xu, Dacheng Tao, and Tianyi Zhou.
\newblock A survey on knowledge distillation of large language models.
\newblock \emph{arXiv preprint arXiv:2402.13116}, 2024.

\bibitem[Yang et~al.(2021)Yang, Wang, Liu, Gao, Ren, Zhang, Wang, Ma, Hua, and Gao]{yang2021towards}
Lingbo Yang, Pan Wang, Chang Liu, Zhanning Gao, Peiran Ren, Xinfeng Zhang, Shanshe Wang, Siwei Ma, Xiansheng Hua, and Wen Gao.
\newblock Towards fine-grained human pose transfer with detail replenishing network.
\newblock \emph{IEEE Transactions on Image Processing}, 30:\penalty0 2422--2435, 2021.

\bibitem[Yang et~al.(2022)Yang, Zhang, Song, Hong, Xu, Zhao, Zhang, Cui, and Yang]{yang2022diffusion}
Ling Yang, Zhilong Zhang, Yang Song, Shenda Hong, Runsheng Xu, Yue Zhao, Wentao Zhang, Bin Cui, and Ming-Hsuan Yang.
\newblock Diffusion models: A comprehensive survey of methods and applications.
\newblock \emph{ACM Computing Surveys}, 2022.

\bibitem[Ye et~al.(2023)Ye, Zhang, Liu, Han, and Yang]{ye2023ip}
Hu Ye, Jun Zhang, Sibo Liu, Xiao Han, and Wei Yang.
\newblock Ip-adapter: Text compatible image prompt adapter for text-to-image diffusion models.
\newblock \emph{arXiv preprint arXiv:2308.06721}, 2023.

\bibitem[Yu et~al.(2022)Yu, Xu, Koh, Luong, Baid, Wang, Vasudevan, Ku, Yang, Ayan, et~al.]{yu2022scaling}
Jiahui Yu, Yuanzhong Xu, Jing~Yu Koh, Thang Luong, Gunjan Baid, Zirui Wang, Vijay Vasudevan, Alexander Ku, Yinfei Yang, Burcu~Karagol Ayan, et~al.
\newblock Scaling autoregressive models for content-rich text-to-image generation.
\newblock \emph{Transactions on Machine Learning Research}, 2022.

\bibitem[Zhang et~al.(2023{\natexlab{a}})Zhang, Chen, Chai, Wu, Lagun, Beeler, and De~la Torre]{zhang2023inclusive}
Cheng Zhang, Xuanbai Chen, Siqi Chai, Henry~Chen Wu, Dmitry Lagun, Thabo Beeler, and Fernando De~la Torre.
\newblock {ITI-GEN}: Inclusive text-to-image generation.
\newblock In \emph{ICCV}, 2023{\natexlab{a}}.

\bibitem[Zhang et~al.(2021)Zhang, Li, Lai, and Yang]{zhang2021pise}
Jinsong Zhang, Kun Li, Yu-Kun Lai, and Jingyu Yang.
\newblock Pise: Person image synthesis and editing with decoupled gan.
\newblock In \emph{CVPR}, pages 7982--7990, 2021.

\bibitem[Zhang et~al.(2023{\natexlab{b}})Zhang, Rao, and Agrawala]{zhang2023adding}
Lvmin Zhang, Anyi Rao, and Maneesh Agrawala.
\newblock Adding conditional control to text-to-image diffusion models.
\newblock In \emph{ICCV}, pages 3836--3847, 2023{\natexlab{b}}.

\bibitem[Zhang et~al.(2022)Zhang, Yang, Lai, and Xie]{zhang2022exploring}
Pengze Zhang, Lingxiao Yang, Jian-Huang Lai, and Xiaohua Xie.
\newblock Exploring dual-task correlation for pose guided person image generation.
\newblock In \emph{CVPR}, pages 7713--7722, 2022.

\bibitem[Zhang et~al.(2023{\natexlab{c}})Zhang, Zhou, Wang, Wang, and Yan]{zhang2023patch}
Shaofeng Zhang, Qiang Zhou, Zhibin Wang, Fan Wang, and Junchi Yan.
\newblock Patch-level contrastive learning via positional query for visual pre-training.
\newblock In \emph{ICML}, 2023{\natexlab{c}}.

\bibitem[Zhang et~al.(2023{\natexlab{d}})Zhang, Zhu, Zhao, and Yan]{zhang2022contextual}
Shaofeng Zhang, Feng Zhu, Rui Zhao, and Junchi Yan.
\newblock Contextual image masking modeling via synergized contrasting without view augmentation for faster and better visual pretraining.
\newblock In \emph{ICLR}, 2023{\natexlab{d}}.

\bibitem[Zhang et~al.(2023{\natexlab{e}})Zhang, Zhu, Zhao, and Yan]{zhang2023patch2}
Shaofeng Zhang, Feng Zhu, Rui Zhao, and Junchi Yan.
\newblock Patch-level contrasting without patch correspondence for accurate and dense contrastive representation learning.
\newblock \emph{ICLR}, 2023{\natexlab{e}}.

\bibitem[Zhou et~al.(2022)Zhou, Yin, Chen, Sun, Gao, and Li]{zhou2022cross}
Xinyue Zhou, Mingyu Yin, Xinyuan Chen, Li Sun, Changxin Gao, and Qingli Li.
\newblock Cross attention based style distribution for controllable person image synthesis.
\newblock In \emph{ECCV}, pages 161--178. Springer, 2022.

\bibitem[Zhu et~al.(2019)Zhu, Pan, Chen, and Yang]{zhu2019dm}
Minfeng Zhu, Pingbo Pan, Wei Chen, and Yi Yang.
\newblock Dm-gan: Dynamic memory generative adversarial networks for text-to-image synthesis.
\newblock In \emph{CVPR}, pages 5802--5810, 2019.

\end{thebibliography}
}
\clearpage

%%%%%%%%%%%%%%% APPENDIX %%%%%%%%%%%%%%%
\appendix
\vspace{1mm}
\begin{center}
{
    \hypersetup{linkcolor=citecolor,linktocpage=true}
    \tableofcontents
}
\end{center}
\vspace{1mm}
\section{Ethical and Social Impacts}

Our work in text-based HIG using the HcP layer, which relies on reference images sourced from publicly available datasets and base models publicly released on the HuggingFace diffusers library \cite{von2022diffusers}, presents various ethical and social considerations.
The primary concern is the potential impact on privacy and data protection. The generation of human images based on text inputs could unintentionally produce likenesses of real individuals, highlighting the need for guidelines to protect individual privacy and meanwhile prevent misuse of personal likenesses. Additionally, the inherited biases in the training datasets and base models can lead to stereotypical images. It's important to continuously monitor and adjust these biases to ensure a fair and inclusive representation of generated human figures.

Furthermore, while our method can enhance representation in digital media by generating diverse and accurate human figures, there is a risk of misuse in creating misleading or harmful content. Establishing ethical guidelines and usage policies is crucial to prevent the creation of deepfakes. Collaborative efforts with various stakeholders are necessary to develop responsible use cases and address potential misuse.
In summary, our approach in tHIG, while offering the potential for creative and inclusive image generation, must be balanced with a commitment to ethical practices, privacy protection, and the promotion of diversity and inclusivity in text-based human figure synthesis.

\section{Cross-Attention Layer Details}

The cross-attention layer within the U-Net architecture of latent diffusion models plays a pivotal role in synthesizing detailed and contextually relevant images. This layer operates by computing a set of queries ($\mathbf{Q}$), keys ($\mathbf{K}$), and values ($\mathbf{V}$) based on the input latent representation 
$\mathbf{z}_{in}$ and the given text-conditioned embeddings $\mathcal{C}$. 

First, the input latent representation $\mathbf{z}_{in} $is transformed into a query matrix $\mathbf{Q}$ using a weight matrix $\mathbf{W}_q$. This process converts the input into the query space:
\begin{equation}
    \mathbf{Q} = \mathbf{W}_q ~ \mathbf{z}_{in} \in \mathbb{R}^{d}~,
\end{equation}

Simultaneously, text-conditioned embeddings $\mathcal{C}$ from CLIP \cite{radford2021learning} text encoder, which embeds textual information, is used to generate the key $\mathbf{K}$ and value $\mathbf{V}$ matrices through their respective weight matrices $\mathbf{W}_k$and $\mathbf{W}_v$:
\begin{equation}
    \begin{aligned}
        \mathbf{K} &= \mathbf{W}_k ~ \mathcal{C} \in \mathbb{R}^{d \times N}~,\\
        \mathbf{V} &= \mathbf{W}_v ~ \mathcal{C} \in \mathbb{R}^{d \times N}~,\\
    \end{aligned}
\end{equation}

The attention mechanism then computes the attention map $\mathcal{M}$ by applying a softmax function to the dot product of $\mathbf{Q}$ and $\mathbf{K}^T$, scaled by the square root of the dimensionality $d$. This step effectively captures the relevance of each element in the context of the input latent representation:
\begin{equation}
    \mathcal{M} = \text{softmax} \left( \frac{\mathbf{Q} \mathbf{K}^T}{\sqrt{d}} \right)~,
    \label{equ:softmax}
\end{equation}

Finally, the output latent representation $\mathbf{z}_{out}$ is obtained by multiplying the value matrix $\mathbf{V}$ with a combined attention map $\hat{\mathcal{M}}$ in Eq.~\ref{equ:2}, which is an enhanced version of $\mathcal{M}$ incorporating the novel Human-centric Prior (HcP) layer introduced in our approach:
\begin{equation}
    \mathbf{z}_{out} = \mathbf{V} \times \hat{\mathcal{M}} \in \mathbb{R}^{d}~,
\end{equation}

This augmented cross-attention mechanism, through $\hat{\mathcal{M}}$, effectively integrates human-centric prior information into the diffusion process, leading to more accurate and detailed human image synthesis in the generated images.

\section{Detailed Experiment Settings}
\label{app: setting}
Following the work of Ju \etal \cite{ju2023humansd}, we train the HcP layer on the ensemble of LAION-Human, and the training set of Human-Art, and test on the validation set of Human-Art.
Note that we only apply the text-based prompt in the validation set for inference evaluation.

\mypara{Human-Art} \cite{ju2023humansd} contains 50k high-quality images with over 123k person instances from 5 natural and 15 artificial scenarios, which are annotated with bounding boxes, key points, self-contact points, and text information for humans represented in both 2D and 3D. It is, therefore, comprehensive and versatile for various downstream tasks.

\begin{figure}[h]
\centering
    \includegraphics[width=\linewidth]{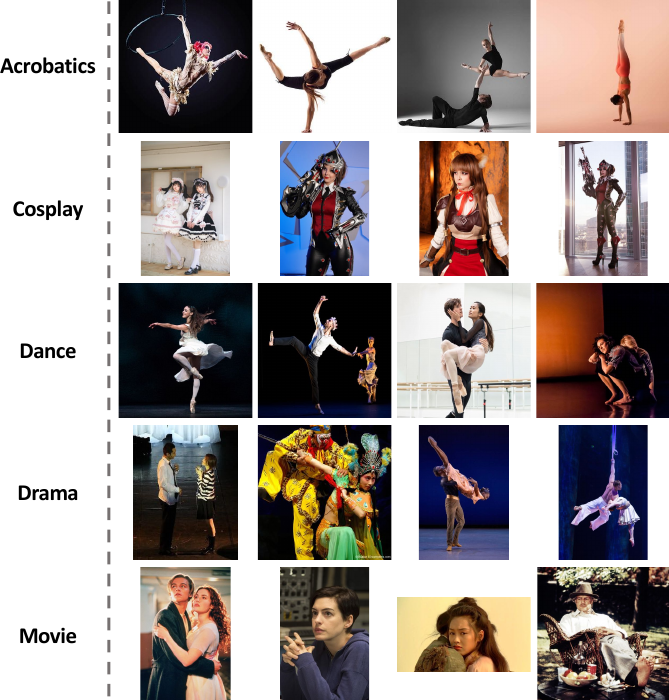}
    \caption{\textbf{Example images from the \textit{realhuman} category in the Human-Art dataset.} Each sub-category (\textit{acrobatics}, \textit{dance}, \textit{drama}, \textit{cosplay}, and \textit{movie}) is represented by four distinct images.}
    \label{fig:example}
\end{figure}

In our study, we specifically focus on the \textit{real human} category of the Human-Art dataset, as it directly aligns with our goal of addressing human image generation challenges. This category encompasses five sub-categories: \textit{acrobatics}, \textit{dance}, \textit{drama}, \textit{cosplay}, and \textit{movie}. These sub-categories offer a diverse range of human actions and poses, providing an ideal context for training the HcP layer to handle complex human structures. Examples are shown in Figure \ref{fig:example}.

\mypara{Laion-Human} \cite{ju2023humansd}.
Ju \etal also constructed a dataset LAION-Human containing large-scale internet images. Specifically, they collected about 1M image-text pairs from LAION-5B \cite{schuhmann2022laion} filtered by the rules of high image quality and high human estimation confidence scores. Superior to ControlNet,
a versatile pose estimator is trained on the Human-Art dataset, which allows for selecting more diverse images such as oil paintings and cartoons. Importantly, LAION-Human contains more diverse human actions and more photorealistic images than data used in ControlNet.

\mypara{Implementation Details}.
Detailed implementation settings of training \& inference stages are listed in Table \ref{tab:impd}. All experiments are performed on 8$\times$ Nvidia Tesla A100 GPUs.
\begin{table}[t]
    \centering
        \caption{List of implementation settings for both training and inference stages.}
    \begin{tabular}{l c  }
    \toprule
      Implementation   & Setting  \\
      \midrule
      Training Noise Schedular &  DDPM \cite{ho2020denoising} \\ 
        Epoch &  10 \\
        Batch size per GPU & 8 \\
        Optimizer &  Adam \\
        Learning rate & 0.0001  \\
        weight decay & 0.01  \\
        Attention map ratio $\gamma$ & 0.1  \\
        loss ratio $\alpha$ & 0.1  \\
        Training timestamp & 1000 \\
        Training image size & 512 $\times$ 512  \\
        Training pose image size & 256 $\times$ 256  \\
        \midrule
         Sampling Noise Schedular &  DDIM \cite{song2020denoising} \\
         Inference step &  50 \\
        guidance scale & 7.5 \\
        Inference image size & 512 $\times$ 512  \\
        \bottomrule
    \end{tabular}

    \label{tab:impd}
\end{table}

\section{Additional Ablations and Analyses}

\subsection{Attention Combination Weight $\gamma$}
\begin{figure}[t]
\centering
    \includegraphics[width=0.9\linewidth]{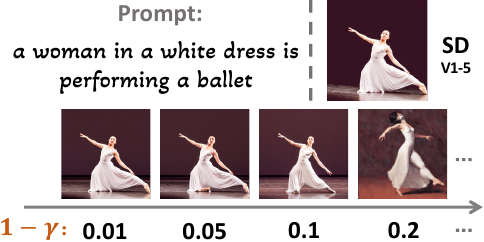}
    \caption{\textbf{Illustration of varying ratios (0.01, 0.05, 0.1, and 0.2) in the attention map combination on image generation.} Images generated for the given prompt at varying attention map combination ratios (0.01, 0.05, 0.1, 0.2).}
    \label{fig:gammavis}
\end{figure}

In this analysis, we focused on the ratio of attention map combination in Eq.~\ref{equ:2} of the main text, as illustrated in Figure \ref{fig:gammavis}. This examination helps us understand the effects of different ratios on the generated images.
At lower ratios of 0.01 and 0.05, the images closely resemble those produced by the standard SD model, indicating that the Human-centric Prior (HcP) layer's adjustments are minimal and maintain the foundational characteristics of the SD outputs.
The most effective correction occurs at a ratio of 0.1, where the HcP layer's influence is well balanced, significantly enhancing the accuracy of human figure generation while maintaining the original style and content of the SD model.
However, a ratio of 0.2 leads to noticeable changes in both the content and style, diverging markedly from the SD model's outputs. Although this ratio corrects the human figures, it also significantly alters the overall image, affecting both the composition and thematic elements.
In conclusion, these observations highlight the importance of an appropriate ratio to achieve a balance between correcting structural inaccuracies and preserving the original style and content of the images. The 0.1 ratio emerges as the most effective, offering an optimal blend of correction and preservation.

\subsection{Learning Process with Pose Image}
\begin{figure}[h]
\centering
    \includegraphics[width=\linewidth]{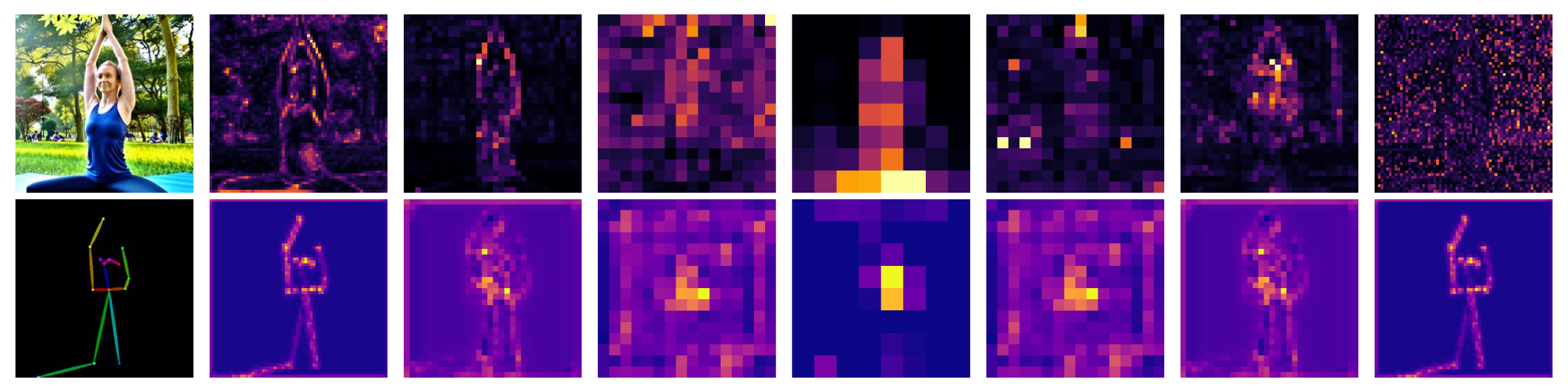}
    \caption{\textbf{Illustration of alignment between layer-specific ResNet features with corresponding scales and combined attention maps in each cross-attention layer.} The ResNet features are extracted from four different scale layers of a ImageNet pre-trained ResNet50 model.}
    \label{fig:pose}
\end{figure}

Figure \ref{fig:pose} provides a visualization of the alignment between layer-specific ResNet features and corresponding scales of the human-centric attention map from the HcP layer. This visualization clearly demonstrates a notable similarity between the layer-specific ResNet features and the cross-attention maps at equivalent scales.
This alignment plays a crucial role in our methodology. By ensuring that the ResNet features, which contain human-centric information, are closely aligned with the corresponding scales of the cross-attention layers, we enhance the model's ability to accurately incorporate human-centric details into the image generation process. This approach not only improves the structural integrity of the generated human figures but also ensures a more contextually accurate representation.

\section{Additional Results}
Due to space limitations, we only provide parts of the results in the main paper. In this section, we will report additional additional results, details, and analyses.

\begin{figure}[t]
\centering
    \includegraphics[width=\linewidth]{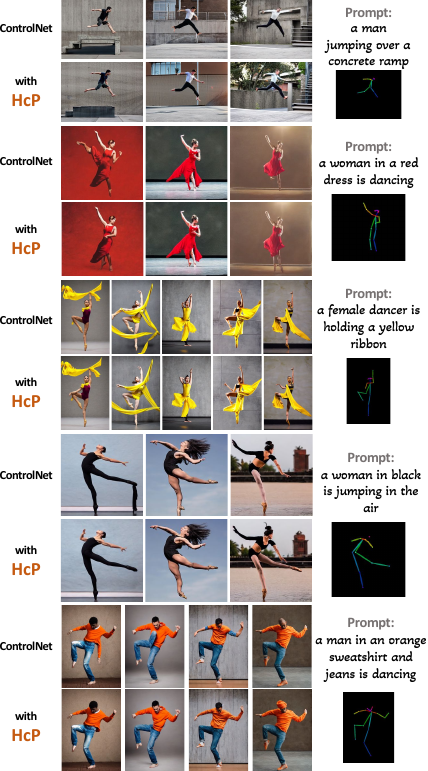}
    \caption{\textbf{Additional comparisons and compatibility with the controllable HIG application.} We selected ControlNet \cite{zhang2023adding} as the basic model for controllable HIG and utilized OpenPose image as the conditional input.}
    \label{fig:control_app}
\end{figure}

\subsection{Human Evaluation Details}
\label{app: human}
We provide the human evaluation setting details here. In the main text, we request participants to evaluate \emph{anatomy quality} (AQ) and \emph{text-image alignment} (TIA) for the prompt-generated image pairs. The former reflects viewers' experience in terms of the anatomical quality of the images. In contrast, the latter reflects viewers' subjective perceptions of text-image consistency between the generated images and corresponding text prompts.
Before rating, we explain the whole procedure of our model and present some example pairs. When displaying these pairs, we explain the definition of AQ and TIA to each participant for a more precise understanding. Pairs that they require to rate are not included in these examples.
Images are displayed in full-screen mode on calibrated 27-inch LED monitors (Dell P2717H). Viewing conditions are in accordance with the guidelines of international standard procedures for multimedia subjective testing~\cite{ITU}. The subjects are all university undergraduate or graduate students with at least two years of experience in image processing, and they claimed to browse images frequently. The percentage of female subjects is about 40\%. All the subjects are aged from 20 to 27 years old.
Before giving the final rating, we allow participants to watch each pair multiple times.

\subsection{Controllable HIG Comparison}
\label{app: controlnet}
We provide additional comparisons with ControlNet \cite{zhang2023adding} as shown in Figure \ref{fig:control_app}.
Note that in most cases, under the control of OpenPose images, ControlNet generates human images with the correct pose. The HcP layer does not interfere with generating human images with correct structures but acts to correct them in cases of errors. This makes the HcP layer an effective tool in preventing the generation of structurally incorrect human images.

\subsection{Human-centric Prior Information Comparison}
\label{app: depth}
\begin{figure}[t]
\centering
    \includegraphics[width=\linewidth]{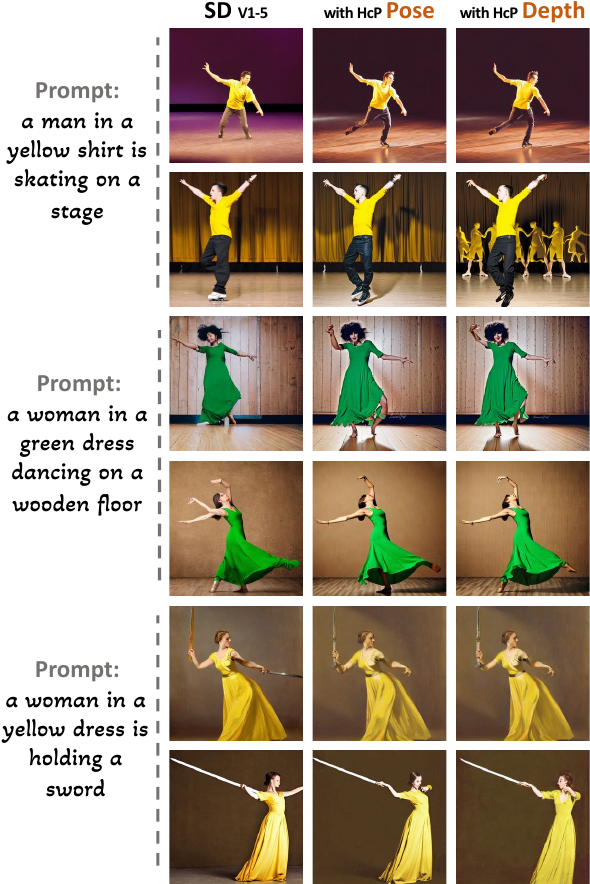}
    \caption{\textbf{Additional comparisons by using different sources of human-centric prior information}. The HcP layer is trained consistently for both Pose and Depth priors to ensure a fair and balanced comparison.}
    \label{fig:depth_app}
\end{figure}

We provide additional results with different sources of human-centric prior information in Figure \ref{fig:depth_app}.

\subsection{Qualitative Results}
\label{app: qualitative}
\begin{figure*}[t]
\centering
    \includegraphics[width=\linewidth]{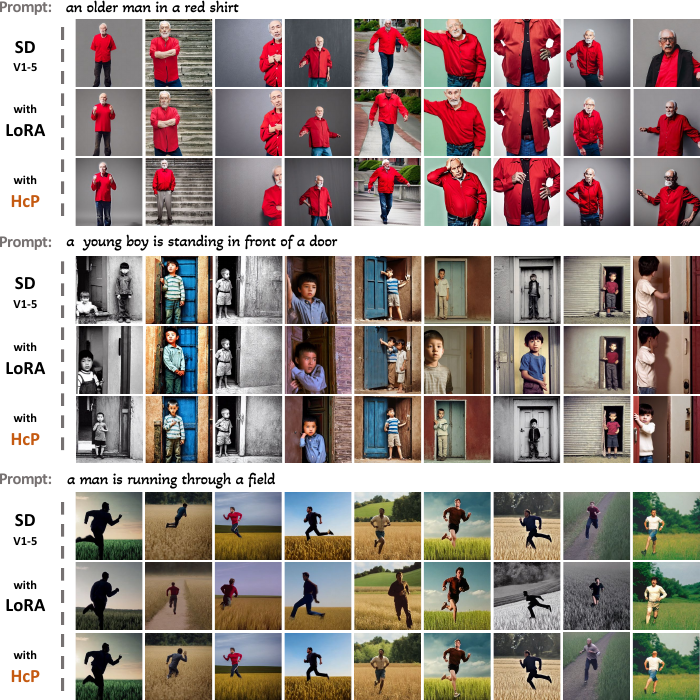}
    \caption{\textbf{Additional qualitative comparison with baseline methods on three example prompts}. We leverage the pre-trained SD v1-5 model for both ``with LoRA" and ``with HcP" models while keeping it frozen.}
    \label{fig:lora_app}
\end{figure*}

\begin{figure*}[t]
\centering
    \includegraphics[width=\linewidth]{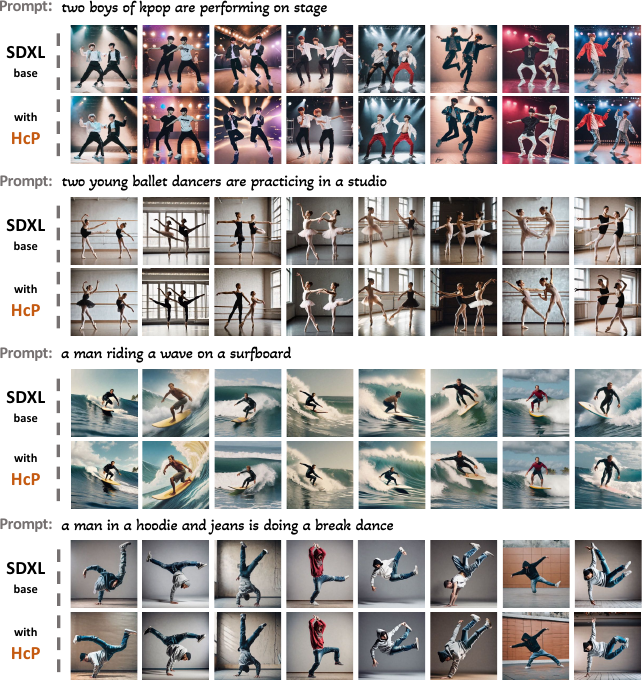}
    \caption{\textbf{Additional results on larger diffusion model (SDXL-base) using HcP layer}. We leverage the pre-trained SDXL-base model for the ``with HcP" model while keeping it frozen.
    }
    \label{fig:sdxl_app}
\end{figure*}

We provide additional qualitative results with baseline methods on three example prompts in Figure \ref{fig:lora_app}, and
we also provide additional large diffusion model (SDXL) results on four example prompts in Figure \ref{fig:sdxl_app}.

\subsection{Failure Cases Analysis}

\begin{figure}[h]
\centering
    \includegraphics[width=\linewidth]{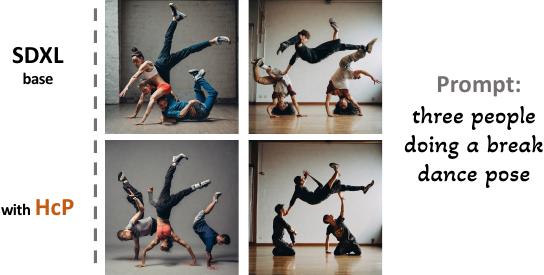}
    \caption{\textbf{Failure cases in complex action scenarios.} Two examples are generated using a pre-trained SDXL-base and an SDXL-base with the integrated HcP layer, in response to a more complex scene text prompt.}
    \label{fig:control_app_fail}
\end{figure}

We examine two instances where the generation of images based on the prompt ``\textit{three people doing a break dance pose}" fell short of expectations in Figure~\ref{fig:control_app_fail}. The main reasons for these limitations are as follows. First, the generation of detailed facial features and limbs is less accurate. This inaccuracy may be due to the limitations of the SDXL-base model itself, particularly when depicting multiple individuals in a complex scene. Second, the intricacy of the `\textit{break dance}' action, combined with the presence of multiple individuals, makes it more challenging to maintain accurate human structure in the generated images.
Despite these challenges, it is noteworthy that the images generated with our HcP layer show improvements in human figure representation compared to those produced by the SDXL-base model. This highlights the HcP layer's effectiveness in enhancing image quality, even in complex scenarios involving detailed movements and multiple subjects.

\section{Futuer work}

Advancing from our present achievements, two crucial areas are highlighted for future development in text-based human image generation:

\mypara{Diverse Data Learning:} To improve the model's capability in handling complex scenarios, we plan to enrich our dataset with more varied and intricate human actions. This will enable continued learning and refinement, enabling better representation of dynamic human interactions.

\mypara{Broader Priors Integration:} We target to incorporate additional human-centric priors simultaneously, such as depth and edge, which will enhance the detail and realism of generated human figures, overcoming the limitations of relying solely on pose information.

\newpage

\newpage

\end{document}